\documentclass[a4paper,fleqn]{cas-dc}

\usepackage{enumitem}
\usepackage{floatrow}
\usepackage{graphicx,subcaption}
\DeclareFloatFont{small}{\small}
\usepackage{float}
\floatstyle{plaintop}
\restylefloat{table}

\usepackage[numbers]{natbib}

\def\tsc#1{\csdef{#1}{\textsc{\lowercase{#1}}\xspace}}
\tsc{WGM}
\tsc{QE}

\begin{document}
\let\WriteBookmarks\relax
\def\floatpagepagefraction{1}
\def\textpagefraction{.001}

\shorttitle{Employing similarity to highlight differences: On the impact of anatomical assumptions in chest X-ray registration methods}

\shortauthors{Berg et al.}  

\title [mode = title]{Employing similarity to highlight differences: On the impact of anatomical assumptions in chest X-ray registration methods}                     

\author[1]{Astrid Berg}[orcid=0000-0002-2300-2661]
\cormark[1]
\fnmark[1]
\ead{berg@vrvis.at}

\credit{Conceptualization, Methodology, Software, Validation, Formal analysis, Data Curation, Writing - Original Draft, Writing - Review \& Editing, Visualization}

\author[2]{Eva Vandersmissen}[orcid=0000-0002-3604-1811]
\ead{eva.vandersmissen@agfa.com}

\credit{Software, Investigation, Supervision, Conceptualization, Visualization, Writing - Review \& Editing}

\author[1]{Maria Wimmer}[orcid=0000-0003-2599-2395]
\ead{mwimmer@vrvis.at}

\credit{Validation, Writing - Review \& Editing, Visualization}

\author[1]{David Major}[orcid=0000-0002-9091-3684]
\ead{major@vrvis.at}

\credit{Validation, Writing - Review \& Editing}

\author[1]{Theresa Neubauer}[orcid=0000-0002-5926-9317]
\ead{tneubauer@vrvis.at}

\credit{Validation, Writing - Review \& Editing, Visualization}

\author[1]{Dimitrios Lenis}[orcid=0000-0002-1563-7683]
\ead{lenis@vrvis.at}

\credit{Validation, Writing - Review \& Editing}

\author[2]{Jeroen Cant}[orcid=0000-0002-0772-252X]
\ead{jeroen.cant@agfa.com}

\credit{Writing - Review \& Editing, Conceptualization}

\author[3, 4]{Annemiek Snoeckx}[orcid=0000-0003-2101-2783]
\ead{Annemiek.Snoeckx@uza.be}

\credit{Conceptualization, Data Curation, Writing - Review \& Editing}

\author[1]{Katja Bühler}[orcid=0000-0002-0362-7998]
\ead{buehler@vrvis.at}

\credit{Conceptualization, Validation, Writing - Review \& Editing, Project administration, Funding acquisition}

\affiliation[1]{organization={VRVis Zentrum f\"{u}r Virtual Reality und Visualisierung Forschungs-GmbH},
	addressline={Donau-City-Straße 11}, 
	city={Vienna},
	postcode={1220}, 
	country={Austria}}

\affiliation[2]{organization={Agfa NV, Radiology Solutions R\&D},
	addressline={Septestraat 27}, 
	postcode={2640}, 
	postcodesep={}, 
	city={Mortsel},
	country={Belgium}}

\affiliation[3]{organization={Department of Radiology, Antwerp University Hospital},
	addressline={Drie Eikenstraat 655}, 
	postcode={2650}, 
	postcodesep={}, 
	city={Edegem},
	country={Belgium}}

\affiliation[4]{organization={Faculty of Medicine and Health Sciences, University of Antwerp},
	addressline={Universiteitsplein 1}, 
	postcode={2610}, 
	postcodesep={}, 
	city={Wilrijk},
	country={Belgium}}

\cortext[cor1]{Corresponding author}

\begin{abstract}	
To facilitate both the detection and the interpretation of findings in chest X-rays, comparison with a previous image of the same patient is very valuable to radiologists. Today, the most common approach for deep learning methods to automatically inspect chest X-rays disregards the patient history and classifies only single images as normal or abnormal. Nevertheless, several methods for assisting in the task of comparison through image registration have been proposed in the past. However, as we illustrate, they tend to miss specific types of pathological changes like cardiomegaly and effusion. Due to assumptions on fixed anatomical structures or their measurements of registration quality, they produce unnaturally deformed warp fields impacting visualization of differences between moving and fixed images.  We aim to overcome these limitations, through a new paradigm based on \textit{individual rib pair segmentation for anatomy penalized registration}. Our method proves to be a natural way to limit the folding percentage of the warp field to $1/6$ of the state of the art while increasing the overlap of ribs by more than $25\%$, implying difference images showing pathological changes overlooked by other methods. We develop an anatomically penalized convolutional multi-stage solution on the National Institutes of Health (NIH) data set, starting from less than 25 fully and 50 partly labeled training images, employing sequential instance memory segmentation with hole dropout, weak labeling, coarse-to-fine refinement and Gaussian mixture model histogram matching. We statistically evaluate the benefits of our method and highlight the limits of currently used metrics for registration of chest X-rays.
\end{abstract}




\maketitle

\section{Introduction}

\begin{figure*}
	\centering
	\includegraphics[width=1.0\linewidth]{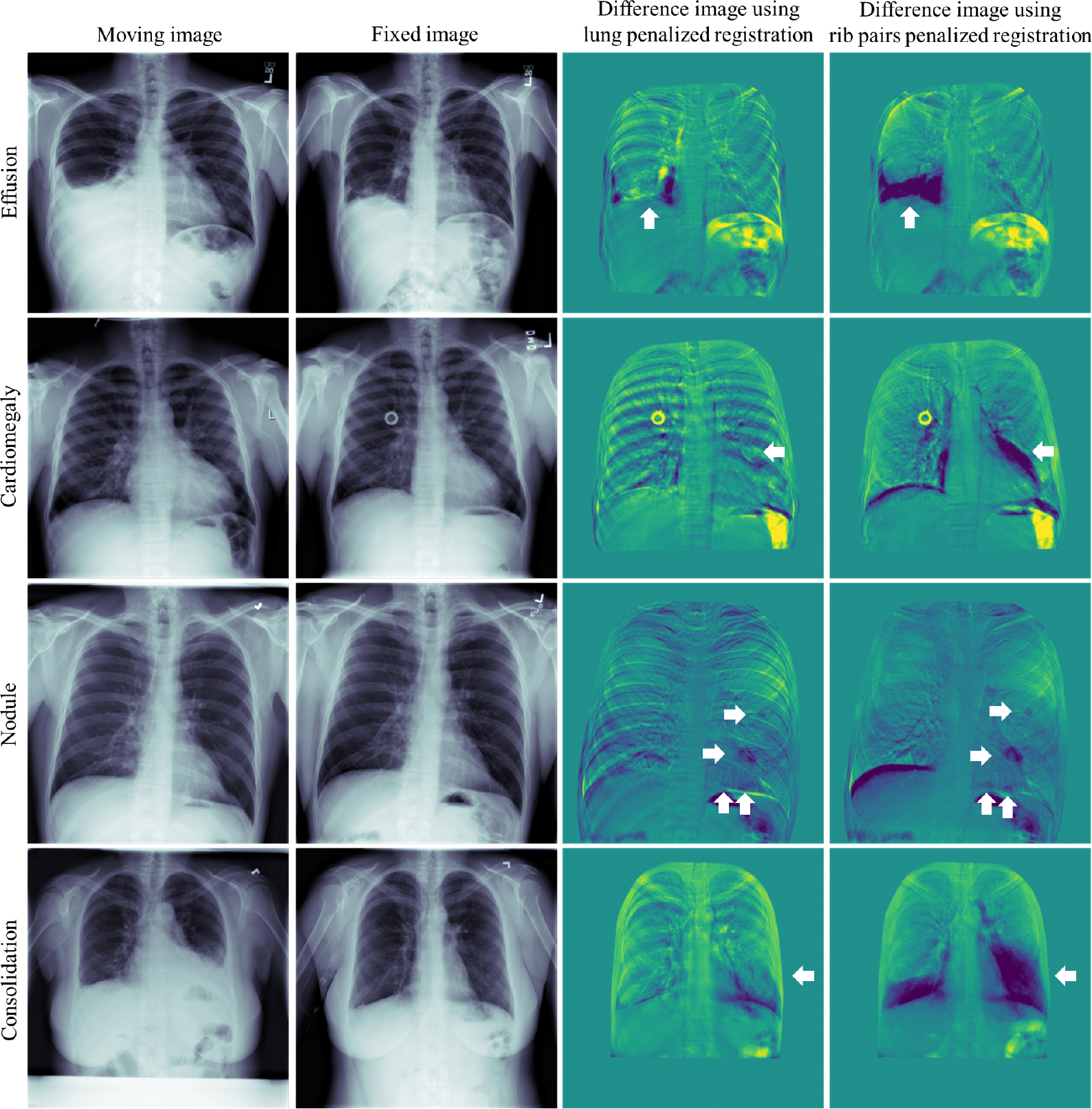}
	\caption{Differences in registration methods for selected pathologies. We illustrate the impact of different types of anatomic penalization on a set of image pairs from the NIH data set, representative for prevalent problems of all SOTA registration methods examined. We trained models for warping the moving (pathological) image to the fixed (healthy) image, optimized for either overlapping lung fields or rib pairs segmentation masks during training. The difference image between the registered images is shown in color, where dark blue correlates with features present in the moving but not in the fixed image (vice versa for yellow). Pathological differences are better visible with our rib pairs penalized registration ($R_{12}$, column $4$) since lung penalized registration methods (e.g. $L_{12}$, column $3$) deform the lung and heart contours which obfuscates these changes. Rib pairs penalization removes spurious differences caused by non-overlapping ribs, while differences in breathing patterns are more pronounced. The visualization can adapt to differently processed images as well as collimation.
}
	\label{fig:lung rib diff}
\end{figure*}

While an abundance of deep learning methods focuses on radiological diagnostics for detecting \textit{specific} types of pathologies on individual images, radiologists screen images for \textit{all} possible types of pathological patterns, like small lesions or changes in heart size and lung opacity. A common diagnostic framework for detecting these patterns on chest X-rays is to compare images of a patient taken at different time steps side by side and to observe changes. Various factors involving image acquisition and the pathology itself complicate this task: images taken with different machines, projections at different angles, different inspiration levels, foreign objects or ribs obstructing the view, as well as pathological changes that move or stay fixed, depending on the chosen reference frame in the 2D projected images. 

Alternative technical approaches in medical image analysis for assisting with this problem have been suggested, like geometric correlation maps \cite{Oh2019}, comparison via Siamese networks \cite{Li2020}, Gated Recurrent Units (GRUs) \cite{Konwer2021} or graph attention \cite{Karawande2022}, but the most commonly used technique remains (deformable) image registration. 
Originally formulated as an optimization problem with high computational costs, learning-based optimization through neural networks has become increasingly popular since the invention of spatial transformers \cite{Jaderberg2015}. Here, a convolutional neural network (CNN) is trained to output a deformation field, that warps the source image to the target image. Correspondence between the warped image and the fixed image is established based on a loss function, which informs the computation of the warp field through back-propagation. At inference time, opposed to conventional methods, optimization is achieved in a single forward pass. Several authors have successfully investigated how to optimally warp the moving/source image to correspond with the fixed/target image based on surrogate measures for chest X-rays with the help of deep learning (see \cite{AlSafadi2021, Fang2020, Mahapatra2019, Mansilla2020}). Although these methods are fast and often surpass classical methods in registration quality, they come with intrinsic challenges, concerning the technical implementations of producing smooth transformations and restricting or limiting folding. 
Apart from technical challenges, there are also remaining ones in the medical domain: How to quantify objectively whether a deformation is medically plausible? How to evaluate if no ground truth for registration is available or when it is tedious to obtain? How to handle large and subtle changes simultaneously?
When comparing registered images, further questions arise: How to quantify or visualize changes? 
We would like to pose an additional question, dealing with the counterintuity of employing image \textit{similarity} to highlight \textit{differences}: Do the current techniques have an inbuilt bias towards missing certain types of pathologies like effusion or cardiomegaly? Our results suggest that the answer is yes. 

\subsection{Contribution} 

The aim of this paper is to illustrate that anatomical and image value biases of different registration methods for chest X-ray registration have an impact on the visualization of the difference image. The difference between these methods can not only be seen in qualitative examples (cf. Figure \ref{fig:lung rib diff}) but can be measured in terms of registration metrics and is indeed statistically significant.  We compare different anatomically penalized registration methods as well as unsupervision and, for the first time, use individual rib pairs as anatomic constraint during training. We aim to counter the impact of anatomic biases of existing lung penalized registration methods, by suggesting a multi-stage deep learning-based registration framework capable of tending to the finer rib details. For this analysis, we need a large data set of patient pairs that is not hand selected for a particular pathology and contains hard cases for registration which is not readily available. Therefore, as a first step, we develop a pipeline for large scale annotation of rib pairs, improving the instance memory approach to sequential segmentation by hole dropout and automatic detection of problematic cases for relabeling, starting from very few labeled images and bootstrapping ourselves up to all posterior-anterior (PA) images of the public NIH data set \cite{Wang2017}. The obtained segmentation masks serve as anatomic penalization in our multi-stage registration framework. We demonstrate that our registration solution can deal with the hard tasks of large pathological lung and heart deformations (cf. Figure \ref{fig:lung rib diff}) while reducing spurious differences stemming from non-overlapping ribs in the visualization (cf. Figure \ref{fig:stages comparison}), where the state of the art (SOTA) is often not applicable or produces misleading results. In contrast to the commonly used lung/heart anatomy or unsupervision, we demonstrate that our method is intrinsically capable of creating plausible deformations even with large changes between images where the rib cage acts as a natural grid for creating transformations that imply a low folding percentage. 

The developed multi-stage architecture proves to be beneficial for different types of anatomic penalization but its benefit becomes especially apparent when measuring rib overlap and folding of the warp field. Our results should stand as a caveat that chest X-ray registration quality should be evaluated task-based, that the answer to the question what should be and what is registered to what is not always intuitive (cf. Table \ref{tab:related work}) and that certain limitations of methods have to be measured differently from the common practice and we provide statistical tests and qualitative visualizations for different types of pathologies to support our hypotheses.
Since registration for chest X-rays is diagnostically used mostly for assessing differences within the same patient, while public data sets mostly consist of a single image per patient, we evaluate the benefits of intra-patient training over inter-patient training for this task. 
To visualize the differences in methods, we employ a Gaussian mixture model histogram matching approach. 

\subsection{Related Work}
\label{sec:related work}

\subsubsection{Deep Learning-Based Registration}
\label{sec:DL reg}

\begin{table*}[h!]
	\caption{Overview of chest X-ray registration methods in the literature. Marked in bold are assumptions on the alignment of lung boundaries, either through the employed penalization method or the evaluation/best model selection process.}
	\label{tab:related work}
	\begin{tabular}{p{0.6cm}p{0.9cm}p{1.6cm}p{1.6cm}p{0.5cm}p{3.3cm}p{2.5cm}p{2.9cm}}
		\hline
		\textbf{Paper}                                                    & \textbf{Method}    & \textbf{Supervision}                             & \textbf{Data}                                                             & \textbf{Pairs} & \textbf{Penalization focus}                                                                                               & \textbf{Target}                                                                                        & \textbf{Evaluation/Best Model Selection}                                                               \\ \hline
		\cite{AlSafadi2021}                             & CNN (single stage) & unsupervised                                     & Montgomery                                                                & inter          & image similarity loss and meta-regularization convolutional layer with radially symmetric, positive semi-definite filters & large deformations                                                                                     & \textbf{DSC of lungs}, percentage of positive Jacobian det., minimum Jacobian det.         \\ \hline
		\cite{Fang2020}, \cite{Guo2021} & CNN \newline (single stage) & unsupervised, \newline weakly supervised \newline pre-registration & private                                                                   & intra          & \textbf{lung based pre-registration} where pairs with large differences are excluded; local cross-correlation                      & improvement of small differences, when already pre-registered through affine and B-spline deformations & visual qualitative evaluation                                                     \\ \hline
		\cite{Mahapatra2019}                             & GAN                & weakly supervised                                & NIH subset                                                                & intra          & \textbf{logarithm of dice overlap of lung masks}                                                                                   & data independent registration/transfer learning between different tasks                                & transfer performance to brain scans                                               \\ \hline
		\cite{Mansilla2020}                              & CNN (single stage) & weakly supervised                                & JSRT, \newline Montgomery, \newline Shenzen & inter          & \textbf{dice overlap of lung/heart segmentation masks and auto encoded features of lung/heart}                                          & improved chest X-ray registration                                                                      & \textbf{DSC, HD, ASSD of left/right lung and heart}                                        \\ \hline \hline
		ours                                                              & CNN (multi stage)  & weakly supervised                                & NIH PA                                                                    & intra          & multi stage penalized overlap of rib cage                                                                                 & large and small deformations                                                                           & DCL, H95L (not used in selection of best model), DCR, H95R, MSE, SSIM, negJAC (cf. Section \ref{sec:metrics}) \\ \hline
		ours                                                              & CNN (multi stage)  & weakly supervised                                & NIH PA                                                                    & intra          & multi stage penalized overlap of rib pairs                                                                                & large and small deformations                                                                           & DCL, H95L (not used in selection of best model), DCR, H95R, MSE, SSIM, negJAC (cf. Section \ref{sec:metrics}) \\ \hline
	\end{tabular}
\end{table*}

Medical image registration has sparked high interest among the deep learning community in recent years (cf. \cite{Fu2020, Haskins2020, Boveiri2020, Chen2020}). Several methods have been investigated to compute and afterwards learn from ground truth (GT) for deformations (cf. for example \cite{Rohe2017, Eppenhof2018, Hoffmann2021}).  
For obtaining plausible deformations, the two main directions followed in the literature are to prevent folding of deformations within the network specification or to impose regularization and penalty terms on the neural network output during training. These methods are often evaluated in terms of the percentage of pixels with negative Jacobian determinant. In the first category, specifically stationary velocity fields (SVFs) are relevant, which are employed for example by \cite{Dalca2018, Shen2019, Niethammer2019}. In \cite{Mok2021}, they additionally employ simultaneous learning of both forward and inverse transformation for unsupervised learning. 
In the second category, we find different methods for penalizing the deformation fields in unsupervised learning: \cite{Zhang2018} also employ an inverse consistent constraint, to ensure that two images are consistently matched in both directions simultaneously. Additionally, they use an anti-folding constraint, penalizing the gradient of the flow at locations with foldings. Contrary to penalizing warping in both directions on the original images, \cite{Kuang2019} and \cite{Kim2021} introduce cycle consistency constraints within their loss functions, penalizing warping of the moving image to the fixed image but also backwarping of the transformed moving image to the original source image. 

Multi-resolution registration has become increasingly popular during the last years, especially for brain MRIs. In \cite{Mok2020}, 
they suggest a multi-stage Laplacian pyramid network, where, at higher resolution stages, the warped image from the previous stage as well as the previous stage velocity field are used as the input. They conclude that SVF parametrizations give better results in terms of Jacobian negative percentage, but displacement field parametrizations give better anatomical structure overlap. A similar concept with joint optimization at multiple stages is also employed by \cite{Li2020a}. Multilevel approaches have also been introduced for inhale-exhale CT registration, for example by \cite{Hering2019} in a weakly lung supervised setting or \cite{He2021} in the unsupervised setting.

Chest X-ray analysis by deep learning-based methods is currently of high scientific and commercial interest (cf. for example \cite{Sogancioglu2021}), in particular also registration. In \cite{Mahapatra2019}, a small set of intra-patient paired images with manually annotated lungs was used for training a generative adversarial network (GAN). Here, pre-registration is done by affine matching and local elastic deformations using B-splines, so only small deformations are targeted. The discriminator network penalizes, among standard terms, also the overlap of lung masks and additionally a cycle consistency loss is used. Small differences are also the subject of \cite{Fang2020} and \cite{Guo2021}, which also start with affine and B-spline registration within segmented lung regions on a private data set, excluding cases with too large deformations. They penalize first and second derivatives of the dense displacement field during CNN training in order to obtain better local matching in an unsupervised setting. 
In \cite{Mansilla2020}, which can be considered the closest to our work, lung anatomy segmentation is used as a prior to inform CNN-based registration during training. These priors are additionally auto encoded, learning low dimensional representations of the segmentation masks.
Regularization of CNN training is also the topic of \cite{AlSafadi2021}, who employ a U-Net type architecture with learned spatial regularization filters which approximate Toeplitz matrices for inter-patient registration. Table \ref{tab:related work} gives an overview of the methods and hints at their respective focus on anatomical penalization of lung overlap either through the method or the evaluation/best model selection process.

\subsubsection{Longitudinal Chest X-Ray Comparison}
\label{sec:longitudinal comp}

Alternatives to registration have been proposed for comparing chest X-rays. In \cite{Santeramo2018}, a modification to the LSTM is developed, taking into account the time lag between consecutive studies. These networks capture the evolution of visual patterns over time, benefiting classification performance of pathologies on the current study. Since in practice it is more relevant to track the development of a lesion over time, regardless of the disease class, \cite{Oh2019} go for a different approach, by building on a two stream feature extractor from the input images. They develop geometric correlation maps, where the output of these two streams are local descriptor feature maps compared by correlation scores for every possible match. A binary classifier is trained to recognize patterns within this map indicative of change. 

Recently, also for COVID-19, several publications focus on X-ray comparison. In \cite{Li2020}, a Siamese network was used to calculate pulmonary disease severity as the Euclidean distance between a pool of healthy images and the current image. In \cite{Konwer2021}, progression of lung infiltrates is predicted in several zones, by taking the previous images and neighboring patches into account via a Gated Recurrent Unit (GRU). Global and local dependencies between anatomical regions are the focus of \cite{Karawande2022}, who employ a graph attention network.

In summary, we find that alternative approaches to leveraging longitudinal information in chest X-rays are mostly developed for improving automatic detection or tracking of the progression for specific pathologies, with the exception of those focusing on abstract feature correlations. Anatomical information used in these methods either focuses on the division of images into lung regions or into anatomical regions, where this information is then also needed during inference. In contrast to registration, which only supports the radiologist in detecting changes, these methods inherently provide pathological classification or localization.

\subsubsection{Rib Segmentation}
\label{sec:rib seg}

While extraction of bone structures from chest X-ray images \cite{Gozes2020} or segmentation of the rib cage through CNN \cite{Wang2020} or generative adversarial network (GAN) architectures \cite{Oliveira2020} is thoroughly investigated in the literature, segmentation and enumeration of individual ribs is less common (cf. Figure \ref{fig:segmasks}). 
In \cite{Berg2016a}, the response of filter banks is used for localizing edges and center lines of ribs. Through geometric constraints and an a priori rib diameter, the contour of the rib is constructed, and modified by a shape model approach \cite{Lorenz2006}. Using labels constructed with this method as a GT, \cite{Mader2018} build a deep learning-based localization method, and employ conditional random fields (CRF) for refining the label assignments. 
In \cite{Wessel2019}, a variant of Mask R-CNN was used to segment individual ribs on chest X-ray images on a data set curated for bone suppression for lung nodule detection \cite{Berg2016}. Anchor boxes for ribs' locations were estimated from the GT data. To this end, they borrow ideas from vertebrae segmentation \cite{Lessmann2019} for sequential processing of ribs.
A new rib benchmark data set with SOTA baseline scores was published in \cite{Nguyen2021}.

\subsubsection{Delimitation}
\label{sec:delim}

In this paper, we are not focused on automatic pathological change detection, but on assisting radiologists in their assessment of change through registration based approaches. 
In contrast to registration methods for chest X-rays using synthetic transformations or weakly supervised approaches, which are mainly focused on lung anatomy, we establish a pipeline for labeling rib pairs and simulate imperfect instance memory through a region-dropout augmentation technique. For the paired rib segmentation we implement simple but effective rule based quality control of these segmentation masks. 
Using this approach, we are able to leverage the registration capabilities of the displacement field formulation while circumventing their limitations. Our registration method works as a multi-stage architecture, where the lower resolution displacement field acts as input to a refinement module on the warp field and the original images, in contrast to previously established multi-stage architectures, concatenating multi-resolution transforms iteratively. 
We show that our method brings regularization properties as inherent benefit to registration, by working as a natural grid that limits folding, as an alternative to anti-folding or cycle consistency constraints. 
This results in better highlighting of large and small pathological differences in the difference image: getting rid of spurious noise stemming from non-overlapping ribs makes small changes better visible while unnatural deformations of the warp field are prevented, illustrating changes in lung and heart boundary missed by other methods. 

\section{Materials and Methods}
\label{sec:methods}

In this section we define our weakly supervised multi-stage registration approach (cf. Figure \ref{fig:architecture_reg}). For the penalization of this model during training, we use rib pairs (cf. Figure \ref{fig:segmasks}), which act as a curved regularization grid. In order to leverage the potential of paired unlabeled chest X-ray data, we employ a sequential rib pairs segmentation framework (cf. Figure \ref{fig:architecture_seg}) in the weak data annotation process of rib pairs (cf. Figure \ref{fig:data_annotation}) with automatic quality checks for relabeling of problematic cases. Through Gaussian mixture model histogram matching we finally visualize the difference image (cf. result images in Figure \ref{fig:lung rib diff}, \ref{fig:result comparison} and \ref{fig:stages comparison}). In the following, we first introduce our data set, then explain the details of our method for registration and visualization and afterwards introduce our automatic annotation process. 

\subsection{Data}
\label{sec:data}

\begin{figure}
	\centering
	\includegraphics[width=1.0\linewidth]{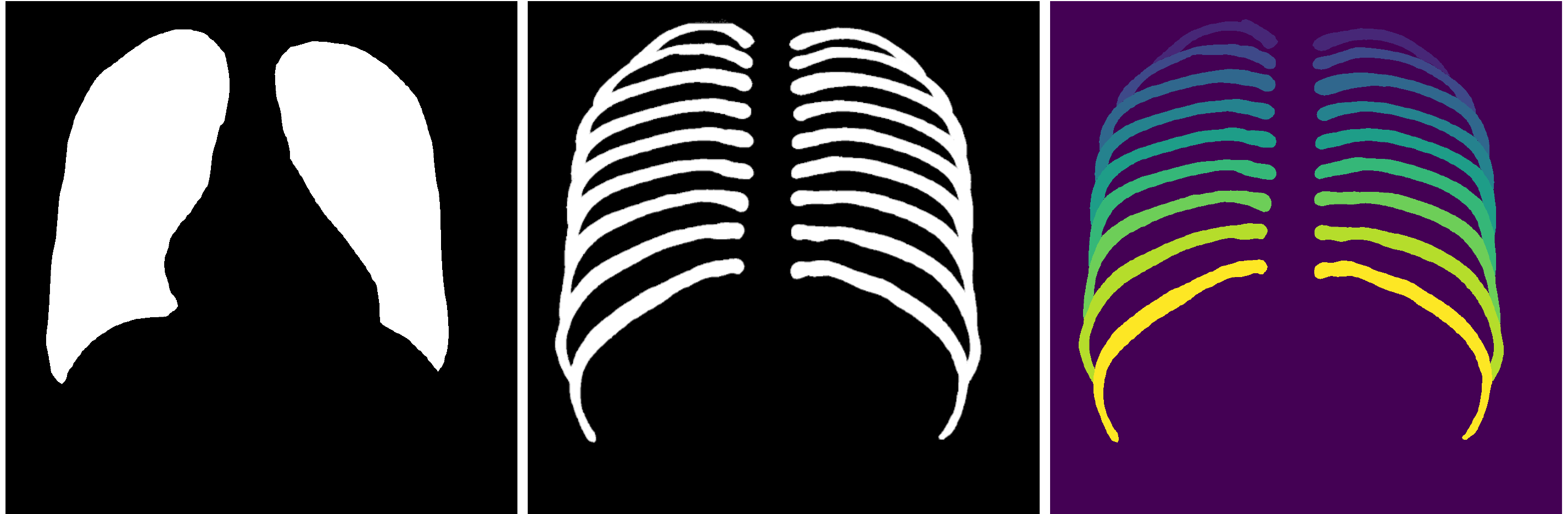}
	\caption{Different segmentation masks used for penalization of weakly supervised models. From left to right: lung, rib cage and rib pairs.}
	\label{fig:segmasks}
\end{figure}

Our experiments are based on the following data and labels (cf. also Figure \ref{fig:segmasks}):

\subsubsection{NIH-PA}
\label{sec:data NIH}

We use the public Chest-XRay14 dataset \cite{Wang2017}, that contains $112.120$ images of patients with $14$ common diseases visible on chest X-rays in the resolution $1024 {\times} 1024$. The advantage of this data set over other public data sets is, that it contains pairs of images of the same patient taken at different times. We filter this data set for the $67.310$ PA images. NLP mined classification labels for pathologies in this data set are available. However, segmentation masks (lungs or ribs) are not available from official sources.

\subsubsection{Rib Annotations}
\label{sec:data annotations}

Manual annotations of ribs were performed on a subset of the NIH-PA data set. For details on the data selection we refer to Section \ref{sec:data split}. We annotate ribs pairwise where a pair, consisting of left and right rib at the same height, shares the same label. We aim for a curved grid provided by these rib pairs, which we enumerate by L2-R2 up to L10-R10. The pair L1-R1 was deliberately not annotated, since it often disappears behind L2-R2 in the $2D$ projection, providing no additional information. Pairs below the $10$th, anterior ribs and clavicles were not labeled. 
Weak rib labels are provided for all NIH-PA images through the annotation pipeline described in Section \ref{sec:methods rib seg} by the models $Seg_1^{(*)}$ and $Seg_n^{(*)}$. Whenever we speak of rib cage annotations as opposed to rib pairs, we convert the multi-label rib pairs to a binary segmentation mask, where all ribs share the label $1$ and the background is $0$.

\subsubsection{Lung Annotations}
\label{sec:data XLSor}

To obtain robust lung labels for a fair comparison of methods, we use the lung segmentation model XLSor by \cite{Tang2019} as the GT for the full NIH-PA data.

\subsection{Registration}
\label{sec:methods registration}

\begin{figure*}[h!]
	\centering
	\includegraphics[trim=0 243 0 2,clip, width=1.0\linewidth]{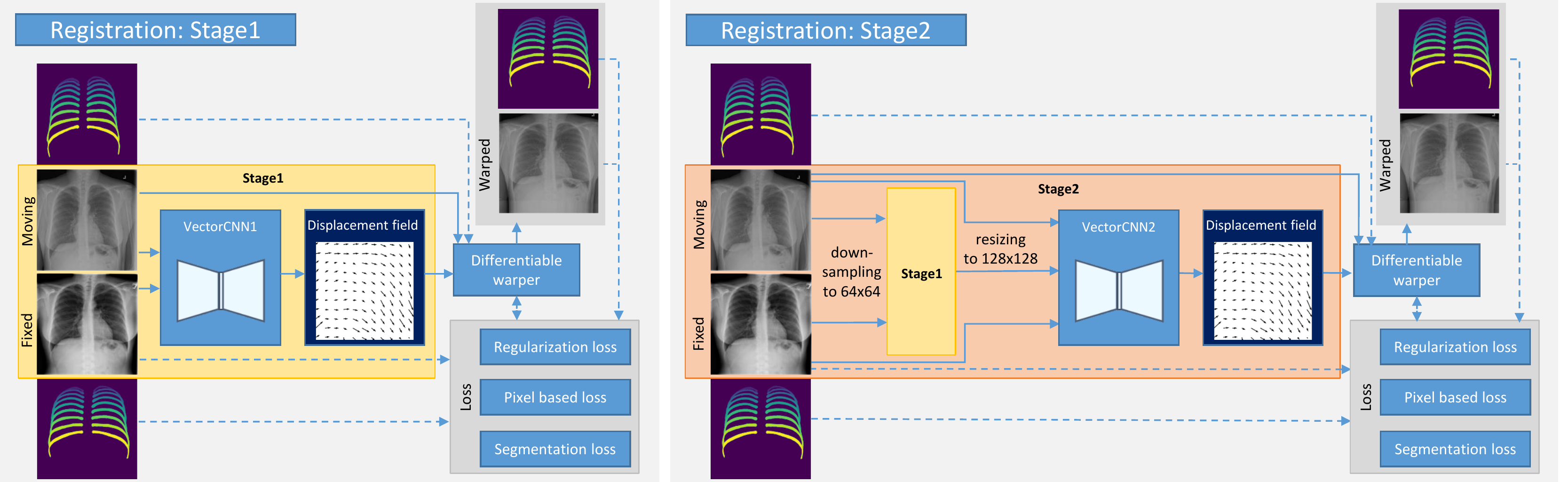}
	\caption{Registration architecture. In stage$1$, VectorCNN1 is trained independently, while in stage$2$, VectorCNN2 and VectorCNN1 are trained simultaneously. Segmentation masks are used for penalization during training (dashed arrows) and not needed for inference. During inference, moving and fixed image are downsized to the resolution of VectorCNN1. The output warp field is upsampled and concatenated with the input images. VectorCNN2 outputs a refined displacement field.}
	\label{fig:architecture_reg}
\end{figure*}

Our multi-stage registration architecture is illustrated in Figure \ref{fig:architecture_reg} and uses rib pairs segmentation masks as anatomic penalization during training. We build a multi-stage registration architecture, starting with the VectorCNN from \cite{Mansilla2020} which serves as the first stage (stage$1$). This is an encoder-decoder type CNN similar to U-Net. Here, a moving and fixed image $M$, $F$ in ${\left[ 0, 1\right]}^{n {\times} n}$, for image input size $n{\times}n$, are concatenated to a single input stack and fed to the network. This network then outputs the displacement field, representing the transformation $T: {\left[ 0, 1\right]}^{n {\times} n} \rightarrow {\left[ 0, 1\right]}^{n {\times} n}$ as a tensor of shape $\left[n, n, 2\right]$. We train this architecture at $64 {\times} 64$ resolution, giving a first estimation of the warp field. While this resolution seems enough to resolve lung boundaries well, it is not enough to attend to the fine structures of rib boundaries. Therefore, we follow a coarse-to-fine refinement approach, where, for the second stage, we concatenate the first stage with a higher input resolution, multi-input VectorCNN in the following way: the input for VectorCNN2 has depth $4$ and is a concatenation of $M$, $F$ at $128{\times}128$ resolution and the bilinearly upsampled displacement field output of VectorCNN1. During stage2, we initialize VectorCNN1 with the weights from stage$1$ training. In contrast to other works (cf. Section \ref{sec:related work}), here stage$2$ acts as a refinement stage directly on the displacement field output. While it is possible to stack this module to register at even higher resolutions in an iterative manner, the two-stage approach followed here is a trade-off between hardware and registration quality. \\

For the penalization of the training, we use weak rib pairs segmentation masks obtained by the method in Section \ref{sec:methods rib seg} and illustrated in Figure \ref{fig:segmasks}. As loss function, we use a combination of established loss functions for registration (cf. for example \cite{Balakrishnan2018} or \cite{Mansilla2020}), i.e. negative normalized cross correlation $ncc$ of the image pairs, total variation $tv$ of the transformation and the categorical cross-entropy $ce$ of the warped segmentation mask $S_{M}$ and the fixed segmentation mask $S_{F}$,

\begin{equation}
\begin{split}
\mathcal{L} (M, F, T) = - ncc (M \circ T, F) & + \lambda_r tv (T) \\
& + \lambda_{seg} ce (S_{M} \circ T , S_{F}). \\
\end{split}
\end{equation}

During inference, the output displacement field is bilinearly upsampled to the original image size of $1024{\times}1024$ and applied to the moving image. For the differentiable warping module, we use the spatial transformer module implementation of \cite{Mansilla2020}.
While this multi-stage architecture was envisioned with the application of chest X-ray registration and anatomic rib pairs penalization in mind, the model can also be successfully applied to other forms of anatomic penalization (cf. Section \ref{sec:training details reg}). 

\subsection{Visualization of differences}
\label{sec:methods visualization}

For visualizing differences, we want to ensure that the method is applicable to different types of machines and the presence of foreign objects and collimation. To this end, we first apply the trained registration network to the moving image at resolution $1024{\times}1024$. 
We restrict the warped image to a ROI, by segmenting the rib cage and computing the convex hull plus a small pixel margin $PM$. Since moving and fixed image might be acquired differently, in order to transfer both images to the same value range, we apply a Gaussian mixture model with $10$ components to the gray-level distribution within the ROI of the warped image. We use the lower boundary $0$, the means found by the mixture model and the maximum of each image to transfer the histograms by piece-wise linear interpolation. Afterwards, we subtract the equalized warped image from the current image to obtain the difference image and clip at $mean \pm 4std$, to ensure that values outside this range are clearly shown as differences. We subtract the mean of the difference image and visualize it with a symmetric color map centered at $0$, which roughly corresponds to "unchanged" in this setting (cf. Figure \ref{fig:result comparison}). Colors on the extreme end of the color map roughly correspond to features which are present in one image but not in the other. Whenever multiple images are shown in the same figure, we ensure that the same ROI and standardization is applied to all images for fair comparison.

\subsection{Rib pairs segmentation}
\label{sec:methods rib seg}

\begin{figure*}
	\centering
	\includegraphics[trim=0 370 0 0,clip, width=1.0\linewidth]{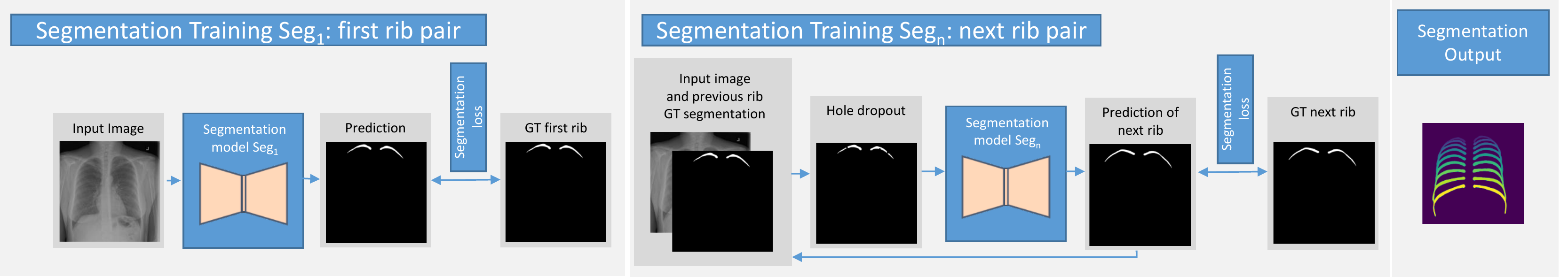}
	\caption{Two step segmentation approach with simulated imperfect ribs by dropout of connected regions. At inference time, the first rib pair is segmented with $Seg_1$ and used as input to $Seg_n$, which is iteratively applied to obtain the segmentation output. Regions of the previous ribs are deleted in the instance memory through dropout augmentation, resulting in an enumeration of ribs that is more stable to noise.}
	\label{fig:architecture_seg}
\end{figure*}

We generate annotations by two supervised models, where one, $Seg_1$, segments the first rib pair and another one, $Seg_n$, the consecutive rib pair, given the input image and the current rib pair GT mask (cf. Figure \ref{fig:architecture_seg}).
We use a U-Net with InceptionResNetV2 \cite{Incep} backbone for both. For $Seg_n$ we use an approach inspired by \cite{Wessel2019, Lessmann2019}, which has been shown to outperform generic multi-class segmentation. During training, if rib pair $i$ should be segmented, we provide the model with the input image and the GT mask of rib pair $i{-}1$ (the \textit{instance memory}) as input stack and compute the loss between the output and GT rib pair $i$. During inference, the model $Seg_1$ is applied to the input image for finding the first rib pair. The first rib pair and the input image are then stacked and provided as input to $Seg_n$, which is applied $8$ times, each time to the input image and segmentation mask $i{-}1$. Since ribs are segmented pairwise, imperfectly segmented rib pairs used as instance memory could consist of multiple connected components confusing the segmentation network. We simulate imperfect segmentation of previous ribs during training by dropout of connected regions on the instance memory input, to get disconnected ribs as a guiding signal.

We develop a pipeline (cf. Figure \ref{fig:data_annotation}) for easy annotation of rib pairs when manual annotations are few. We start by training models $Seg_1^{(1)}$ and $Seg_n^{(1)}$ on few manually labeled examples and use them to annotate further examples weakly. Masks are then automatically quality judged, surviving annotations are incorporated in the training of further models $Seg_1^{(2)}$ and $Seg_n^{(2)}$ with weak labels. During this process, we judge the quality of masks automatically, where the implementation details are explained in Section \ref{sec:quality selection seg}: \\

 Q1 Does a rib pair have more than $2$ components?

 Q2 Does a rib pair consist of only one rib? 

 Q3 Is the symmetry of rib sizes disturbed?

 Q4 Is the symmetry of rib heights disturbed? \\
 
These quality criteria are quite conservative and enable us to quickly find mislabeled hard cases. We relabel those images iteratively, by manually correcting the first wrongly segmented rib pair and applying the sequential segmentation models again with the corrected instance memory. The rib masks are finally used as supervision for models $Seg_1^{(*)}$ and $Seg_n^{(*)}$ which provide anatomic masks for the registration.

\section{Experimental Setup}
\label{sec:experimental setup}

\begin{table*}
	\begin{tabular}{|l|l|ll|}
		\hline
		& \textbf{segmentation}          & \multicolumn{2}{l|}{\textbf{registration}}                                            \\ \hline
		\multirow{6}{*}{\textbf{training}} & \textit{seg-train-strong}                 & \multicolumn{2}{l|}{\multirow{3}{*}{\textit{reg-train-weak}}}                                  \\
		& $\subset$ \textit{seg-train-strong-first} & \multicolumn{2}{l|}{}                                                                 \\
		& $\subset$ \textit{seg-train-weak}  & \multicolumn{2}{l|}{}                                                                 \\ \cline{2-4} 
		& \textit{seg-val-strong}                & \multicolumn{1}{l|}{\multirow{3}{*}{\textit{reg-val-weak}}} & \multirow{3}{*}{\textit{reg-val-strong}} \\
		& $\subset$ \textit{seg-val-strong-first}   & \multicolumn{1}{l|}{}                                &                                \\
		& $\subset$ \textit{seg-val-weak}       & \multicolumn{1}{l|}{}                                &                                \\ \hline
		\textbf{evaluation}                & \textit{seg-test-strong}                & \multicolumn{1}{l|}{\textit{reg-test-unlab}}                 & \textit{reg-test-strong}                 \\ \hline
	\end{tabular}
	\caption{Data set splits. Each cell in the table is a disjoint data set. In order to prevent data leakage between tasks, patient ids do not overlap between sets (seg=segmentation, reg=registration, strong=strong annotations, weak=weak annotations, unlab=unlabeled data, first=first rib pair annotated).
	}
	\label{tab:datasets}
\end{table*}

\subsection{Train, Validation, Test Split}
\label{sec:data split}

For evaluation of segmentation and registration, we decided on a splitting strategy, preventing patient id leakage between train, valid and test sets and between tasks (cf. Table \ref{tab:datasets} for an overview of data set splits described in the following). For random splits we use the percentage of $70$ train / $15$ validation / $15$ test. 

For segmentation, we randomly select $500$ of the NIH-PA data set, such that no patient is selected twice. We split these $500$ images into $350$ train (\textit{seg-train-weak}) / $75$ validation (\textit{seg-val-weak}) / $75$ test (\textit{seg-test-strong}). For the evaluation of segmentation, the $75$ images in test-strong-seg were fully annotated with rib pairs and lung labels.
For the first stage of segmentation training with expert labels, we randomly select $5\%$ of train and validation set for a full segmentation annotation (\textit{seg-train-strong} and \textit{seg-val-strong}) and $5\%$ more for annotation of only the first rib pair (\textit{seg-train-strong-first} and \textit{seg-val-strong-first}), i.e., we have $21$ strongly annotated ($17$ train and $4$ validation) images for training of the sequential rib segmentation and $42$ ($34$ train and $8$ validation) images for the first rib pair segmentation.
For the second stage of segmentation training, we weakly annotate the $350$ training images in \textit{seg-train-weak} and the $75$ validation images in \textit{seg-val-weak} with weak and automatic quality checked labels from the models from the stage before. We manually corrected $98$ first rib images that did not survive automatic quality checks.

For the registration data split, we first exclude the patient ids of the $500$ segmentation images from the full NIH-PA data set. In the following, we describe the random selection of $5$ disjoint sets of image pairs: a weakly annotated train set \textit{reg-train-weak}, for weakly supervised registration training, a weakly annotated validation set \textit{reg-val-weak} for hyperparameter tuning (early stopping), a strongly annotated validation set \textit{reg-val-strong} for hyperparameter search on supervised metrics, and finally, a strongly annotated test set \textit{reg-test-strong} for evaluation of supervised metrics and an unlabeled test set \textit{reg-test-unlab} for evaluation of unsupervised metrics.
To create stratified sets for evaluation of registration across difficult pathologies, we use the NLP mined labels of NIH-PA. We exclude images with combinations of pathologies appearing less than $4$ times in the whole data set and chose the sets \textit{reg-val-strong} and \textit{reg-test-strong} stratified according to the remaining NLP label combinations. To this end, $50$ strongly rib labeled image pairs for validation set (\textit{reg-val-strong}) and test set (\textit{reg-test-strong}) are selected each, where the first image is chosen according to the stratified distribution and the second image is a random image of the same patient. Since most images in the NIH-PA data set are labeled "No Finding", we limit their occurrence to at most $10\%$ to truly assess the registration quality on hard cases. After subtracting these data sets, the remaining images are again split into train (\textit{reg-train-weak}) / test (\textit{reg-test-unlab}) / validation (\textit{reg-val-weak}) sets. 

\subsection{Evaluation Metrics}
\label{sec:metrics}

\begin{figure*}[ht]
	\centering
	\begin{subfigure}[b]{0.48\linewidth} 
		\centering
		\includegraphics[width=\linewidth]{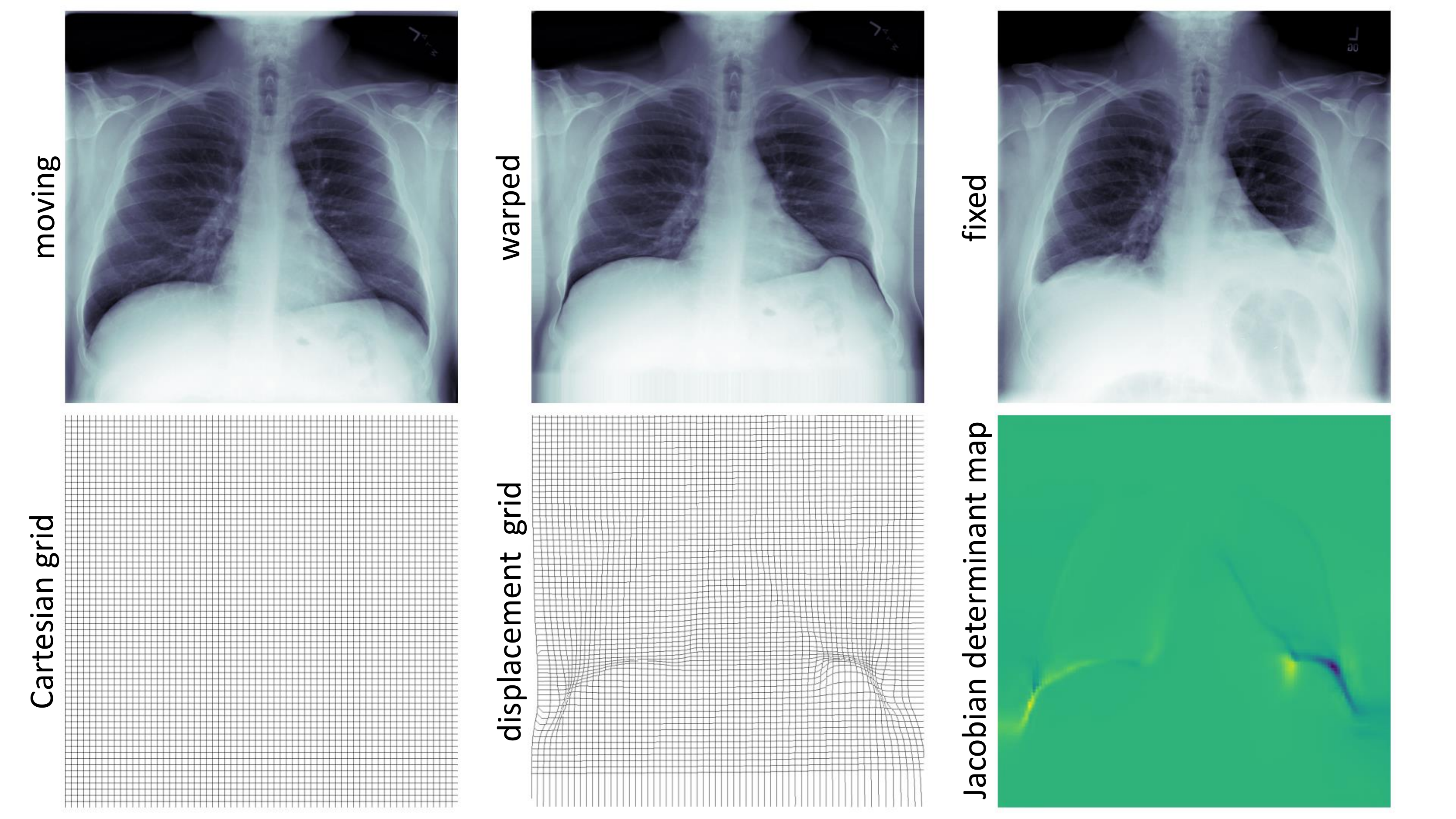}
		\caption{negJAC computation.}
	\end{subfigure}
	\begin{subfigure}[b]{0.48\linewidth} 
		\centering
		\includegraphics[width=\linewidth]{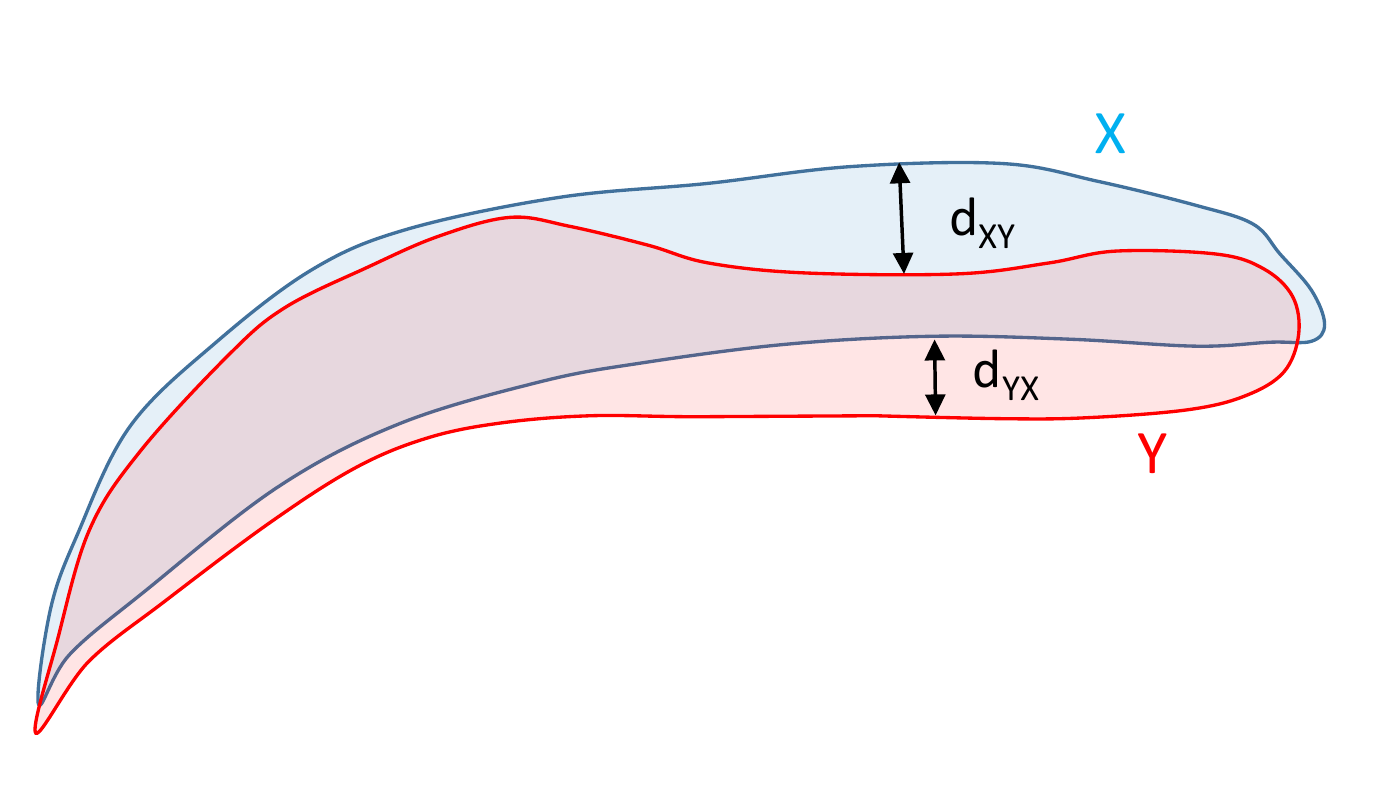}
		\caption{Hausdorff distance computation.}
	\end{subfigure}  
	\caption{Illustration of negJAC and Hausdorff distance computation. (a) For negJAC, the moving image is warped to the fixed image and a Cartesian grid (illustrated in downsampled resolution) is deformed to a displacement grid. Locally, the determinant of the Jacobian for the deformation of each grid cell is computed. Pixels having a negative Jacobian determinant are displayed in dark blue. For negJAC, we calculate the percentage of these pixels in the whole image. (b) The Hausdorff distance is computed by taking the maximum of $d_{XY}$ and $d_{YX}$, where $d_{XY}$ and $d_{YX}$ are the oriented distances between two segmentation masks $X$ and $Y$. This is done for individual rib pairs (here illustrated for one rib) and averaged over all rib pairs.} 
	\label{fig:metrics} 
\end{figure*} 

We use standard evaluation metrics for assessing our methods. For sets $X$ and $Y$, the dice overlap score \textbf{D} is
\begin{equation*}
	\textbf{D} = \frac{2 * | X \cap Y |}{|X| + |Y|},
\end{equation*}
and the Hausdorff distance \textbf{H} is computed by
\begin{align*}
	\textbf{H}(X,Y) = max\{&d_{XY}, d_{YX}\} \\
		   = max\{ &max_{x \in X} min_{y \in Y} d(x,y), \\
		   &max_{y \in Y} min_{x \in X} d(x,y)\}.
\end{align*}
For the $95$th percentile of the Hausdorff distances \textbf{H95}, the distance calculation is based on the $95$th percentile of the distances between boundary points in $X$ and $Y$.

For evaluating, whether lungs in two images are anatomically overlapping, we use: \textbf{DCL} = mean of dice overlap scores of each lung field between two segmentation masks; \textbf{H95L} = mean of $95$th percentile of Hausdorff distances of lung fields between two segmentation masks. Additionally, for evaluating whether ribs are anatomically overlapping in two images, we use: \textbf{DCR} = mean over all dice overlap scores of individual rib pairs between segmentation masks - for segmentation between model output and GT mask, for registration between segmentation masks of warped and fixed image; \textbf{H95R} = mean of $95$th percentile of the Hausdorff distances of individual rib pairs between segmentation masks. 

For assessing unnatural deformations of an image, we use the percentage of negative Jacobian determinants \textbf{negJAC}. Here, we first compute an image, where the scalar value at each pixel is the Jacobian determinant of the warp field, i.e. of the displacement transformation $T$ at this location, computed by 
\begin{align*}
\mathrm{det}[ dT/dx ] = \mathrm{det}[ I + du/dx].
\end{align*}
The metric \textbf{negJAC} is then calculated as the percentage of pixels in this image having a negative Jacobian determinant (cf. Figure \ref{fig:metrics}). 

For assessing pixel level differences between methods in an unsupervised manner, we use: the mean squared error \textbf{MSE} between warped and fixed image, which we define by
\begin{align*}
 	\textbf{MSE} = \sum_{i=1}^{N} \frac{1}{N} | x_i - y_i |^2,
\end{align*}
where $x_i$ and $y_i$ are pixels at the same image position and $i$ ranges over all image pixels. We also use the structural similarity index measure \textbf{SSIM} between warped and fixed image, computed between windows $x$ and $y$ by
 
\begin{align*}
 	\textbf{SSIM}(x,y) = \frac{(2\mu_x\mu_y + c_1)(2\sigma_{xy} + c_2)}{(\mu_x^2 + \mu_y^2 + c_1)(\sigma_x^2 + \sigma_y^2 + c_2)},
\end{align*}
where $\mu_x$ is the pixel sample mean of $x$ (resp. $y$), $\sigma_x^2$ is the variance of $x$ (resp. y), $\sigma_{xy}$ is the covariance of $x$ and $y$ and $c_1$ and $c_2$ are stability constants. Here, we use the default scikit-image implementation for these parameters. We want to point out that we do not claim that \textbf{MSE} and \textbf{SSIM} are metrics particularly suited for evaluating registration quality but we still report them for illustrating differences that would otherwise go unnoticed. 

\subsection{Training and implementation details}

\subsubsection{Segmentation}
\label{sec:training details seg}

We trained the model $Seg_1$, starting from ImageNet weights, using soft dice loss on one channel output. We used train and validation batch size of $4$, Adam optimizer with initial learning rate $1\mathrm{e}{-04}$ and cosine annealing and image resolution of $1024 {\times} 1024$. We pre-processed the input to be image wise in the range $[-1, 1]$ and trained for maximum $500$ epochs with a patience of $80$ epochs in the early stopping callback. For augmentations, we used the albumentations library \cite{Buslaev2020}, in particular: GridDistortion, ElasticTransform, ShiftScaleRotate and HorizontalFlip. For the model $Seg_n$, we used the same parameters as above, with the exception of using a two channel output, one for the foreground and one for the background segmentation of each rib pair. We used a softmax output function and used the average of all channel's dice losses as the total loss function. The $3$-channel input is the stacked image ($2$ times) and the mask of the previous rib pair as instance memory. As additional augmentation, we applied CoarseDropout (max holes: $100$, max height: $20$, max width: $20$, fill value: $0$) on the instance memory input, to simulate imperfect segmentation of the previous pair.

\subsubsection{Quality Selection}
\label{sec:quality selection seg}

\begin{figure}
	\centering
	\includegraphics[width=1\linewidth]{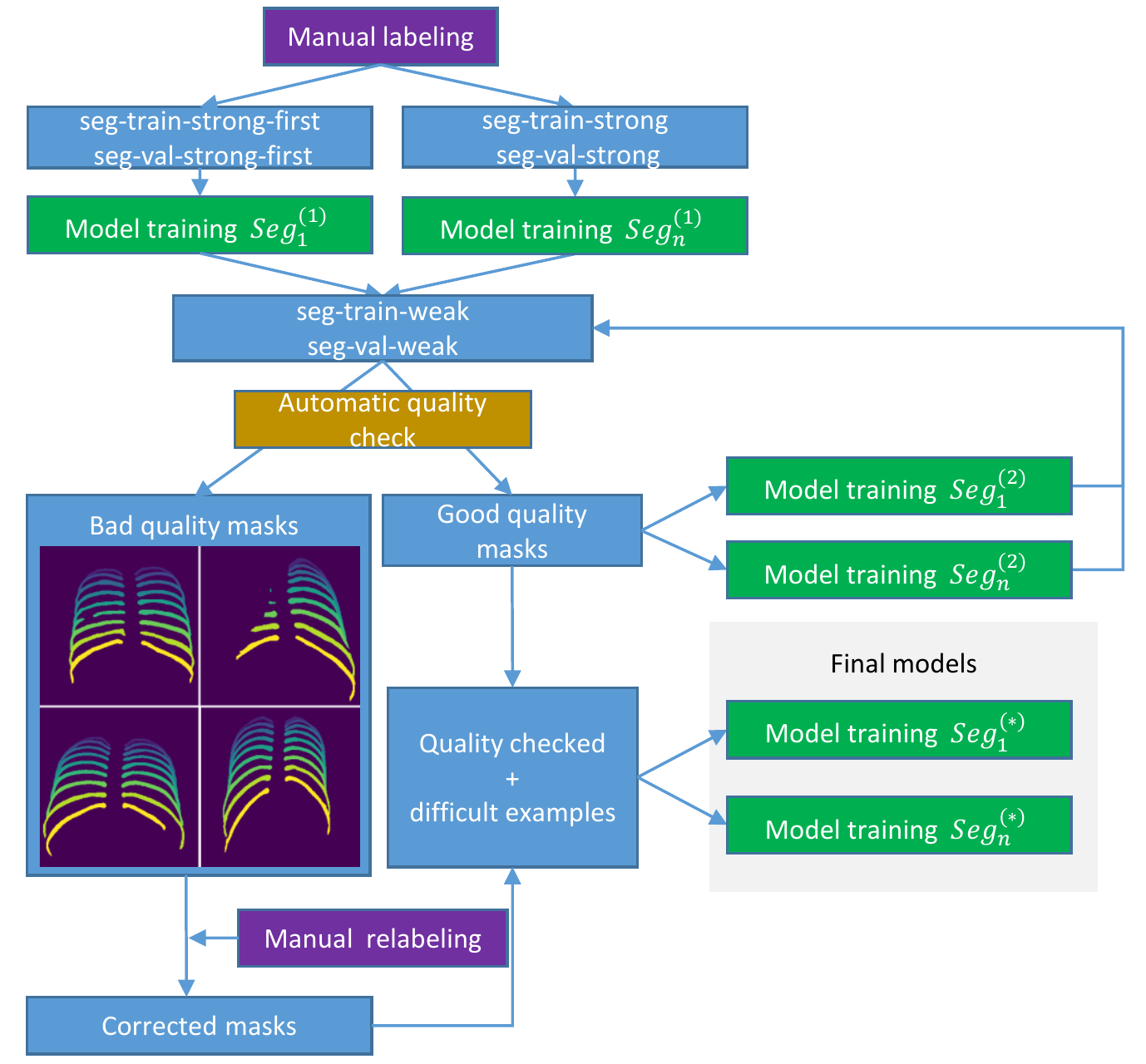}
	\caption{Annotation pipeline for rib pairs: Segmentation models $Seg_1^{(1)}$ and  $Seg_n^{(1)}$ for finding the first and sequential rib pairs are trained on strongly annotated images. These models are used to label images weakly, further models are trained on these weak labels. Quality of labels is automatically judged and if not sufficient, manually corrected. Bad quality masks: upper left: too many components (Q1), upper right: too little components (Q2), lower left: mismatched rib sizes (Q3), lower right: mismatched heights (Q4).}
	\label{fig:data_annotation}
\end{figure}

We explain the computational details of the quality rules described in Section \ref{sec:methods rib seg} and visualized in Figure \ref{fig:data_annotation}.
For Q1, we count the connected components of the segmented rib pair and if there are more than $2$, we determine a lower pixel threshold, where small mislabeled patches are tolerated. We calculated mean and std for the pixel count of the smallest rib pair L2-R2 on every GT mask and set the threshold $T_{Q1} = mean - 2.5 std$. For Q2, we determine whether a rib pair consists of only a single connected component. In Q3, we assess, whether one rib is more than a percentage threshold of $T_{Q3}$ larger than the other rib of this rib pair by setting the threshold to the maximum observed difference percentage in the GT images. For Q4, we assess, whether the highest points of both ribs (i.e. the minimal x-coordinate of the segmentation mask) are more than $T_{Q4}$ pixels apart in height by comparing to the maximum observed pixel distance in GT images. 

Based on the GT masks, we computed $T_{Q1} = 300$, $T_{Q2}$ as a binary flag, $T_{Q3}=30$ and $T_{Q4}=50$. For the selection of $T_{Q3}$ and $T_{Q4}$, we compared the largest connected component for each rib in a pair, ignoring small disconnected patches discarded by Q1. Images that did not survive these criteria were manually corrected. We want to mention that, once the first rib pair is corrected and used as instance memory in $Seg_n$ for segmentation of consecutive ribs, most masks obtained in this way satisfy also the other quality criteria, providing an easy way to get further automatic quality checked labels by simply correcting the first rib pair.

\subsubsection{Registration}
\label{sec:training details reg}

Our registration training works in two stages (cf. Figure \ref{fig:architecture_reg}), where stage$1$ is trained first and the trained weights are then used as initialization in the multi-stage architecture. 
Additionally to the SOTA baseline models, we trained the following architectures based on the RegNet architecture as described in Section \ref{sec:methods registration} for comparing impacts of stage$1$ and stage$2$ training, inter- vs. intra-patient training and for comparing the impact of different anatomies: $L_{1}$, $L_{12}$: stage$1$ and stage$2$ lung penalized intra-patient trained models; 
$RC_{1}$, $RC_{12}$: stage$1$ and stage$2$ binary rib cage penalized intra-patient trained models; 
$R_{1}$, $R_{12}$: stage$1$ and stage$2$ rip pairs penalized intra-patient trained models;  
$R_{ir,1}$: the equivalent stage$1$ inter-patient trained model.

For inter-patient training, an epoch consists of one pass through all randomly shuffled training images and a randomly paired image. All images are first scaled to $\left[0, 1\right]$. We train for 200 epochs with early stopping patience of 20 epochs. For intra-patient training, we show pairs of images of the same patient in all possible combinations to the model. We adjust the number of epochs in order to make sure that approximately the same number of images is shown to the models during training. We train for 40 epochs and 5 epochs of early stopping patience. We use the same hyperparameters for intra- and inter-patient models which we found through hyperparameter grid search for inter-patient stage$1$ and stage$2$ training on the validation set. We base stage$2$ training off the best model found by the grid search in terms of DCR for rib models and DCL for lung models. We found the best hyperparameters to be a learning rate $lr = 1\mathrm{e}{-03}$, $\lambda_{seg} = 3$ for cross entropy loss and $\lambda_{r} = 6\mathrm{e}{-05}$ for total variation loss for stage$1$ and $\lambda_{seg} = 3$ and $\lambda_{r} = 3\mathrm{e}{-05}$ for total variation loss in stage$2$ models for each anatomic penalization. 

\subsubsection{Baseline models}
\label{sec:baseline}

We compare our methods against the SOTA in chest X-ray registration, namely AC-RegNet, the baseline lung penalized registration of \cite{Mansilla2020} and Meta-Reg., the baseline unsupervised method of \cite{AlSafadi2021}, both outperforming classical and deep learning-based methods previously established. These methods stand as prototypical examples for our assessments, together with the ablation single and multistage lung based models described in Section \ref{sec:training details reg}. 
Parameters for the baseline AC-RegNet were chosen according to hyperparameter search for the stage$1$ lung model, with $\lambda_{ae}$ being $1/10$th of the weight of $\lambda_{seg}$ for the parameters specified in \cite{Mansilla2020}. For the baseline network Meta-Reg., we trained the half-U-Net implementation of \cite{AlSafadi2021} with their parameters for chest X-ray registration: regularizing filters of size $13 {\times} 13$, learning rate of $7.5\mathrm{e}{-04}$, and $\lambda=0.05$, $\rho_1 = 0.1$, $\rho_2 = 0.001$.

\subsubsection{Visualization}
\label{sec:implementation vis}
We restrict the visualization to the area of the lung, by segmenting the rib cage and computing the convex hull plus a margin $PM$ of $20px$. We point out that images in the NIH data set are already resized, therefore this is a parameter adapted to the visualization of this specific data set and should ideally be tailored to the original image acquisition resolution and the size and positioning of the patient.

\section{Results and Discussion}
\label{sec:results}

\begin{figure*}
	\centering
	\includegraphics[width=0.98\linewidth,trim=1 0 2 0,clip]{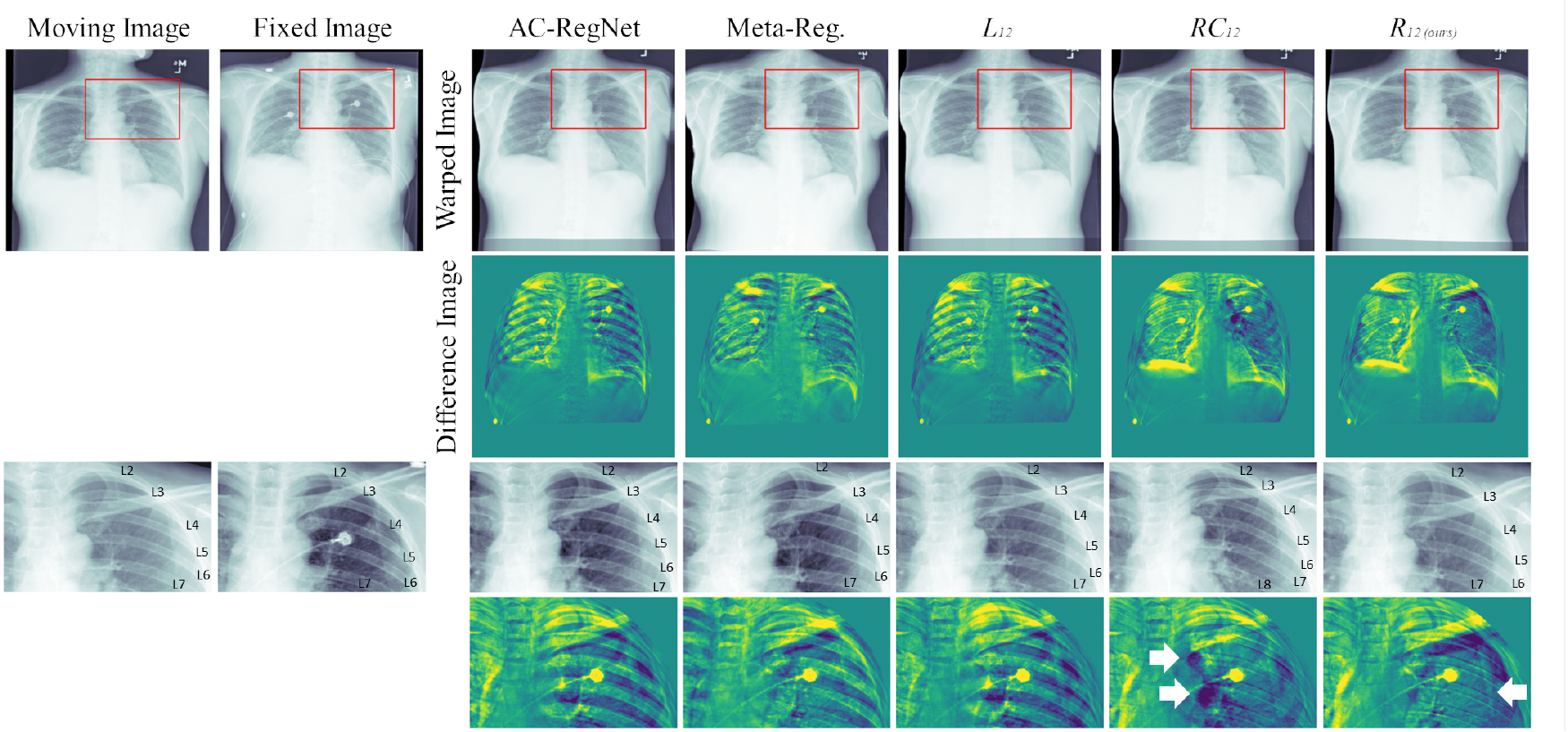}
	\caption{Registration results comparison. While lung penalized (AC-RegNet, $L_{12}$) and unsupervised (Meta-Reg.) methods introduce noise in the difference image in form of a "zebra crossing" pattern by not overlapping ribs, the rib cage penalized method $RC_{12}$ misaligns ribs and wrongly suggests pathological differences (dark spots highlighted by white arrows, last row). Only rib pairs penalization $R_{12}$ correctly aligns ribs and makes fine details visible (medial border of the scapula highlighted by white arrow, last row). Less distraction makes difference in density between left and right lung better visible (second row), while non-overlapping clavicles and diaphragm boundary create additional differences.}
	\label{fig:result comparison}
\end{figure*}

\begin{table*}[h!]
	\begin{tabular}{llllllll}
		\toprule
		Name 			& DCR $\uparrow$ 			   			& H95R $\downarrow$  						& DCL $\uparrow$ 			    		& H95L $\downarrow$  					& MSE $\downarrow$	 			  		 & SSIM $\uparrow$ 			   		 & negJAC $\downarrow$					\\ \midrule
		AC-RegNet \cite{Mansilla2020}	  	& 0.5 $\pm$ 0.2  			& 33.4  $\pm$ 38.9				& \textbf{0.967	$\pm$ 0.015}& \textbf{17.11 $\pm$ 14.49}& 0.0167 $\pm$ 0.022  		 & 0.708 $\pm$ 0.058  		 & 0.0018 $\pm$ 0.005			\\ 
		Meta-Reg. \cite{AlSafadi2021}	  	& 0.383 $\pm$ 0.2 			& 57.21 $\pm$ 52.3				& 0.911	$\pm$ 0.088   		& 51.27 $\pm$ 54.82			& \textbf{0.0085 $\pm$ 0.008}& \textbf{0.724  $\pm$ 0.05}& 0.011 $\pm$ 0.018			\\ \midrule
		$L_{1}$  	& 0.503	$\pm$ 0.2  			& 33.09 $\pm$ 38.76				& 0.965	$\pm$ 0.017			& 17.12 $\pm$ 17.23& 0.0167 $\pm$ 0.022  		 & 0.708 $\pm$ 0.059  		 & 0.0019	$\pm$ 0.004		\\ 	
		$L_{12}$	& 0.503 $\pm$ 0.2  			& 32.31 $\pm$ 37.03				& 0.967	$\pm$ 0.016& 17.58	$\pm$ 15.7			& 0.0185 $\pm$ 0.023	  	 & 0.704 $\pm$ 0.059   		 & 0.0007	$\pm$ 0.002		\\	\midrule		
		$RC_{1}$	& 0.747	$\pm$ 0.19 			& 22.28 $\pm$ 37.35				& 0.932	$\pm$ 0.041			& 40.14	$\pm$ 31.24			& 0.0192 $\pm$ 0.021  		 & 0.707 $\pm$ 0.059  		 & 0.0018	$\pm$ 0.004		\\ 
		$RC_{12}$	& 0.777	$\pm$ 0.17 			& 20.03 $\pm$ 36.25				& 0.929	$\pm$ 0.041			& 40.18	$\pm$ 29.13			& 0.0205 $\pm$ 0.023  		 & 0.706 $\pm$ 0.06  		 & 0.0014	$\pm$ 0.005		\\ \midrule
		$R_{1}$	    & 0.764	$\pm$ 0.16 			& 19.54 $\pm$ 35.42				& 0.932	$\pm$ 0.043			& 41.89	$\pm$ 37.34			& 0.0195 $\pm$ 0.023  		 & 0.707 $\pm$ 0.059  		 & 0.0008	$\pm$ 0.003		\\ 
		$R_{12}$	& \textbf{0.791 $\pm$ 0.16} & \textbf{18.09 $\pm$ 35.13}	& 0.929	$\pm$ 0.041			& 41.19 $\pm$ 29.37			& 0.02  $\pm$ 0.022	  		 & 0.706 $\pm$ 0.06   		 & \textbf{0.0003 $\pm$ 0.001}	\\
		\bottomrule
	\end{tabular}
	\caption{Scores of best intra-patient models. These models were trained on the full training set \textit{reg-train-weak}, evaluation scores are computed on the test set \textit{reg-test-strong} of $50$ strongly annotated images. Different models perform well in different metrics, but our rib pairs penalized model $R_{12}$ not only significantly improves DCR but also reduces negJAC.}
	\label{tab:results}
\end{table*}

\begin{table*}[h!]
	\begin{tabular}{llllllll}
		\toprule
		Name 			& DCR $\uparrow$ 			   			& H95R $\downarrow$ 						& DCL $\uparrow$ 			    		& H95L $\downarrow$ 					& MSE $\downarrow$	 			  		 & SSIM $\uparrow$ 			   		 & negJAC $\downarrow$					\\ \midrule
		AC-RegNet \cite{Mansilla2020}		& 0.494 $\pm$ 0.19			& 33.4 $\pm$ 37.75				& \textbf{0.965 $\pm$ 0.018}& \textbf{18.56 $\pm$ 17.54}& 0.020 $\pm$ 0.019			 & 0.704 $\pm$ 0.064		 & 0.0034 $\pm$ 0.009		\\
		Meta-Reg. \cite{AlSafadi2021}		& 0.302 $\pm$ 0.18			& 61.36 $\pm$ 48.33				& 0.895 $\pm$ 0.086			& 58.05 $\pm$ 52.67			& \textbf{0.0127 $\pm$ 0.010}& \textbf{0.71  $\pm$ 0.06} & 0.0118 $\pm$ 0.017		\\ \midrule
		$L_{1}$		& 0.501 $\pm$ 0.19			& 33.14 $\pm$ 37.69				& 0.964 $\pm$ 0.018			& 19.08 $\pm$ 18.04			& 0.0203 $\pm$ 0.019		 & 0.705 $\pm$ 0.064		 & 0.004  $\pm$ 0.01		\\
		$L_{12}$	& 0.502 $\pm$ 0.2			& 32.28 $\pm$ 37.24				& 0.965 $\pm$ 0.019			& 19.32 $\pm$ 17.5			& 0.0222 $\pm$ 0.020		 & 0.7   $\pm$ 0.065		 & 0.0022 $\pm$ 0.007		\\ \midrule
		$RC_{1}$	& 0.712 $\pm$ 0.2			& 25.04 $\pm$ 38.15				& 0.931 $\pm$ 0.041			& 40.27 $\pm$ 32.03			& 0.0229 $\pm$ 0.020		 & 0.703 $\pm$ 0.065		 & 0.0037 $\pm$ 0.009		\\
		$RC_{12}$	& 0.759 $\pm$ 0.19			& 21.29 $\pm$ 37.77				& 0.930 $\pm$ 0.043			& 40.87 $\pm$ 31.22			& 0.0244 $\pm$ 0.021		 & 0.702 $\pm$ 0.066		 & 0.0021 $\pm$ 0.006		\\ \midrule
		$R_{1}$		& 0.738 $\pm$ 0.17			& 21.59 $\pm$ 36.23				& 0.931 $\pm$ 0.04			& 40.32 $\pm$ 29.23			& 0.0234 $\pm$ 0.021		 & 0.701 $\pm$ 0.065		 & 0.0028 $\pm$ 0.008		\\
		$R_{12}$	& \textbf{0.775 $\pm$ 0.18} & \textbf{19.81 $\pm$ 36.66}	& 0.930 $\pm$ 0.041			& 40.37 $\pm$ 28.93			& 0.0249 $\pm$ 0.022		 & 0.701 $\pm$ 0.067		 & \textbf{0.0017 $\pm$ 0.006}\\
		\bottomrule
	\end{tabular}
	\caption{Average scores of intra-patient models trained on $10$ independent train splits. Metrics needing GT (DCR, H95R, DCL, H95L) were computed on \textit{reg-test-strong}, unsupervised metrics (MSE, SSIM, negJAC) on $10$ independent test set splits of \textit{reg-test-weak} and averaged. Training with less data results in DCR reduction compared to Table \ref{tab:results}.}
	\label{tab:results splits}
\end{table*}

\begin{table*}[h!]
	\begin{tabular}{llllllll}
		\toprule
		Name 			& DCR $\uparrow$ 			   			& H95R $\downarrow$ 						& DCL $\uparrow$ 			    		& H95L $\downarrow$ 					& MSE $\downarrow$	 			  		 & SSIM $\uparrow$ 			   		 & negJAC $\downarrow$						\\ \midrule
		$R_{1}$		& \textbf{0.738 $\pm$ 0.17}	& \textbf{21.59 $\pm$ 36.23}	& 0.931 $\pm$ 0.04			& 40.32 $\pm$ 29.23			& 0.0234 $\pm$ 0.021		 & \textbf{0.701 $\pm$ 0.065}& \textbf{0.0028 $\pm$ 0.008}	\\
		$R_{ir,1}$		& 0.686 $\pm$ 0.19			& 26.42 $\pm$ 36.76				& \textbf{0.931 $\pm$ 0.039}& \textbf{40.07 $\pm$ 29.99}& \textbf{0.0224 $\pm$ 0.02} & 0.700 $\pm$ 0.063		 & 0.0035 $\pm$ 0.01			\\
		\bottomrule
	\end{tabular}
	\caption{Average scores of intra-patient and inter-patient models trained on $10$ independent train splits (supervised metrics on \textit{reg-test-strong}, unsupervised metrics averages on \textit{reg-test-weak} splits). Intra-patient training significantly increases DCR and reduces negJAC, while DCL is not significantly different.}
	\label{tab:intra vs inter}
\end{table*}

\begin{figure*}[h!]
	\centering
	\includegraphics[width=1.0\linewidth]{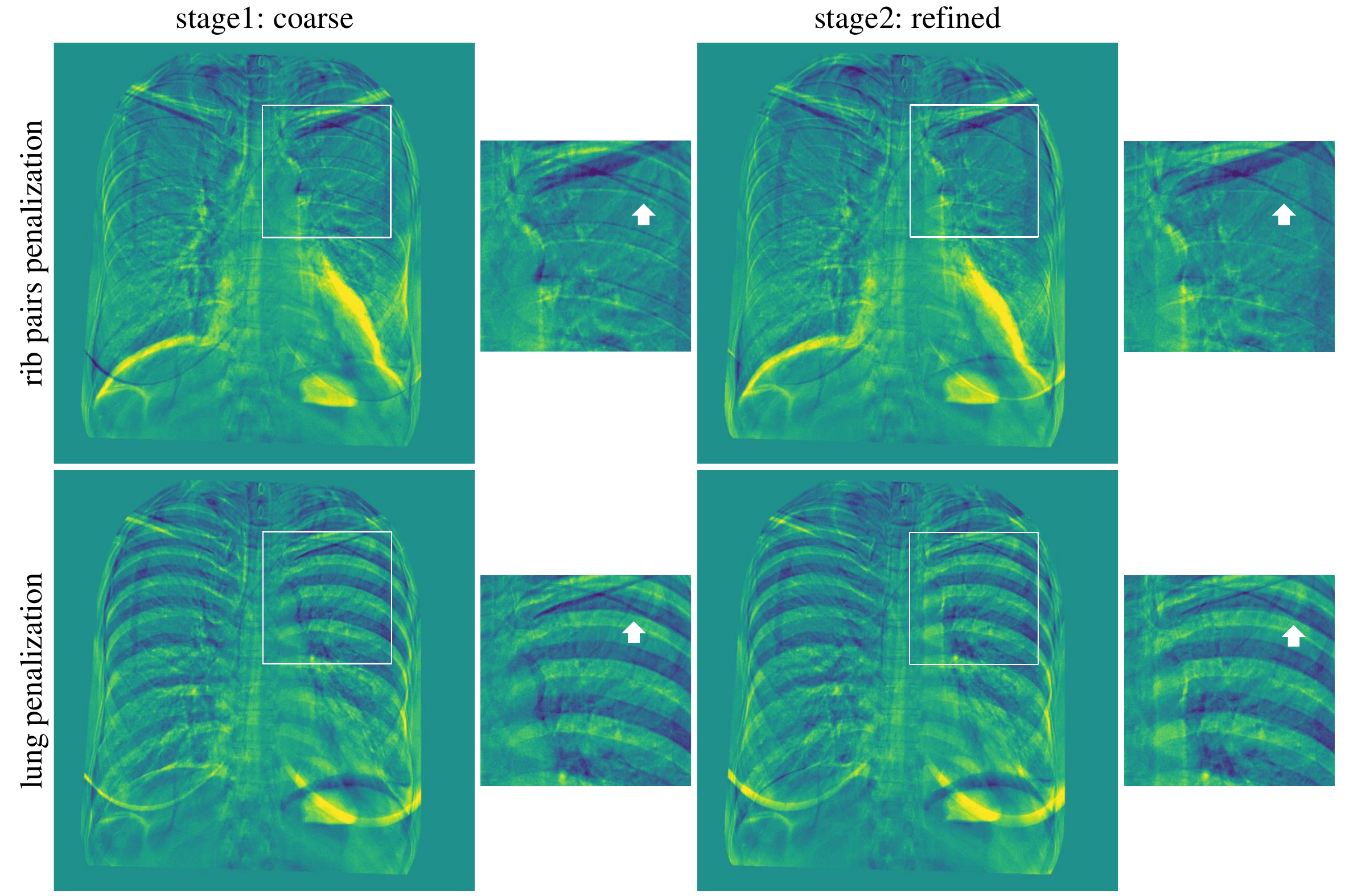}
	\caption{Impact of the refinement stage demonstrated on the difference image. The multi-stage architecture using rib pairs penalization results in statistically significant improved overlap of rib boundaries. This method does not force lung boundaries to overlap, therefore differences in lung/heart shape become visible. Rib pairs penalization reduces spurious differences caused by non-overlapping ribs. Additionally, fine details like the thin catheter in the magnified image section become more visible.}
	\label{fig:stages comparison}
\end{figure*}

We start by providing the technical details of our statistical analysis setup in Section \ref{sec:statistical analysis} and the evaluation scores in Section \ref{sec:scores}. In the further sections we statistically test and analyze the meaning of our findings in detail.

\subsection{Statistical Significance Analysis Setup}
\label{sec:statistical analysis}

For the selection of statistical tests, we follow the recommendations of \cite{demsar06a} to investigate differences between models with non-parametric tests and corrections for multiple testing.
For the statistical analysis of our segmentation methods, we compared DCR of the trained models on \textit{seg-test-strong} over all images with the Friedmann test ($\mathcal{F}$) and the Nemenyi post-hoc test ($\mathcal{N}$) with $\alpha = 0.005$. For registration, we perform different tests for unsupervised metrics (negJAC, MSE, SSIM) and metrics needing GT for evaluation (DCR, DCL). We compute independent scores for unsupervised metrics to account for randomness in initialization and dependence on train and test set. To this end, we split \textit{reg-train-weak} and \textit{reg-test-unlab} each in $10$ parts, train models on the $10$ disjoint splits and evaluate on the disjoint test set splits. We use $\mathcal{F}$ with $\alpha = 0.005$ and apply $\mathcal{N}$ with $\alpha = 0.05$ for investigating significant rank differences. For comparing our results to other methods we use the Wilcoxon signed rank test with Bonferroni correction ($\mathcal{W}$) for the number of pairwise comparisons between our method and all others. For pairwise tests we use a common significance level of $\alpha = 0.05$ before correction to make multiple testing correction possible, even though most individual tests satisfy $p < 0.005$. For supervised metrics we compute scores on the common test set \textit{reg-test-strong}. We train all models with the same parameters as on the whole data set with the exception that we set the maximal epochs and patience to $10$ times the number of the original data set. In terms of evaluation scores, we observe the same trends on the average scores over $10$ splits in Table \ref{tab:results splits} as in Table \ref{tab:results}.

\subsection{Model Evaluation Scores}
\label{sec:scores}

In this section, we want to provide the scores of our model evaluations and demonstrate that differently penalized models benefit from evaluation with different types of metrics. Models were trained according to the specifications in Section \ref{sec:training details seg} and \ref{sec:training details reg}. 

All registration models were trained on \textit{reg-train-weak}, weakly annotated with our best segmentation models $Seg_1^{(*)}$ and $Seg_n^{(*)}$. On \textit{reg-test-strong}, our method $R_{12}$ achieves the best DCR of $0.791 \pm 0.16$, the best H95R of $18.09 \pm 35.13$ and the best negJAC of $0.0003 \pm 0.001$. As illustrated in Table \ref{tab:results}, the best DCL of $0.967 \pm 0.015$ and H95L of $17.11 \pm 14.49$ are achieved by AC-RegNet, while the best MSE of $0.0085 \pm 0.008$ and the best SSIM of $0.724 \pm 0.05$ are achieved by Meta-Reg. We want to point out that models trained with a chosen anatomic penalization consistently outperform those trained with a different penalization throughout the grid search in terms of the metric measuring this anatomic overlap, e.g., lung penalized models with weaker penalization of lung overlap do not learn to align ribs as good as rib penalized models. For inter- vs. intra-patient training comparison on stage$1$ (cf. Table \ref{tab:intra vs inter}), we achieved an average of $0.738 \pm 0.17$ DCR on \textit{reg-test-strong} over the $10$ train splits for the intra-patient trained model $R_{1}$ and an average of $0.686 \pm 0.19$ DCR for inter-patient $R_{ir,1}$. 

For the annotation pipeline, we evaluate the benefit of hole dropout and incorporating weak labels over the established instance memory approach from the literature \cite{Wessel2019, Lessmann2019}. Our best segmentation model for the first rib pair $Seg_1^{(*)}$, trained on \textit{seg-train-weak} achieved a dice overlap of the first rib of $0.9 \pm 0.11$ on \textit{seg-test-strong}. We based the sequential rib segmentation on the common segmentation of this first rib and evaluated DCR on \textit{seg-test-strong}. Here, we achieved $0.9 \pm 0.13$ DCR for $Seg_n^{(1)}$ when training only on strongly labeled images \textit{seg-train-strong}, $0.91 \pm 0.14$ DCR for the same model trained with hole dropout and $0.94 \pm 0.14$ for our best model $Seg_n^{(*)}$ trained on the larger weakly labeled set \textit{seg-train-weak}.

\subsection{Rib penalization vs. other methods}

We first want to investigate the impact of rib penalization over other penalization methods when only training a single stage. To this end, we ask the following questions: 
\begin{enumerate}[label=(\alph*)]
	\item \label{a} Is the benefit of individual rib penalization over other method in terms of rib overlap statistically significant?
	\item \label{b} Is the benefit of individual rib penalization for reducing folding of the warp field statistically significant?
	\item \label{c} Is there a true benefit of using rib pairs over rib cage penalization, as Figure \ref{fig:result comparison} would suggest?
\end{enumerate}
Secondly, we want to investigate lung penalization: 
\begin{enumerate}[label=(\alph*), resume]
	\item \label{d} Is the benefit of lung penalized methods over other methods in terms of overlapping lung fields statistically significant?
\end{enumerate}

For \ref{a}, we want to analyze whether the higher DCR of rib pairs models $RC_{1}$ and $R_{1}$ compared to the other single stage models observed in Table \ref{tab:results splits} is statistically significant. $\mathcal{F}$ finds significant differences in DCR and negJAC within all single stage models. $\mathcal{N}$ finds the differences in DCR between both rib methods and baselines significant. With $\mathcal{W}$, we confirm these findings and also find DCR differences between $RC_{1}$ and $R_{1}$ over all other stage$1$ methods significant, which answers \ref{a} affirmatively. 

For \ref{b} and \ref{c}, $\mathcal{N}$ finds the difference in negJAC to be significant between rib pairs method $R_{1}$ and all other methods except AC-RegNet. With $\mathcal{W}$, we find negJAC differences between $R_{1}$ and all other stage$1$ models to be significant, in particular also to AC-RegNet and the full cage model $RC_{1}$. This answers \ref{b} affirmatively. 

For \ref{c}, with $\mathcal{W}$ we also find evidence that the rib pairs penalized method $R_{1}$ outperforms the rib cage method $RC_{1}$ in terms of DCR. This gives a strong argument for using individual ribs as penalization as opposed to the binary rib cage and answering question \ref{c} affirmatively. All these points together statistically support the claim that rib methods are indeed superior in preventing the warp field from collapsing while better overlapping rib boundaries. 

We also found statistical evidence that lung-based penalization ($L_{1}$ and AC-RegNet) outperforms other methods in terms of DCL by $\mathcal{N}$ and $\mathcal{W}$, which answers \ref{d} affirmatively. We did, however, not find the differences between these weakly supervised lung penalized methods to be significant. \\

As an additional observation that was not rigorously statistically tested due to time constraints, we observed that models trained on the full data set achieve noticeably better DCR scores than those trained on $1/10$ of the training data, while for DCL the differences are not that prominent (cf. Table \ref{tab:results} and Table \ref{tab:results splits}). This might give an indication that collecting more data does provide increased value for the more challenging rib penalization compared to lung penalization. 

\subsection{Weakly supervised vs. unsupervised methods}

We statistically investigate the strengths and weaknesses of weakly supervised and unsupervised methods. With $\mathcal{W}$, we find that weakly supervised models outperform the unsupervised model Meta-Reg. not only in the metric measuring the anatomical overlap they were trained on but also in overlap tasks for unobserved structures (DCR for lung-based models $L_{1}$ and AC-RegNet and DCL for rib penalized models $RC_{1}$ and $R_{1}$). We hypothesized, that contrary to this, Meta-Reg. would outperform weakly supervised models on pixel difference metrics. $\mathcal{W}$ finds significant differences in MSE between Meta-Reg. and all other methods, differences in SSIM are not significant.
Visually, we observed this focus on pixel differences of unsupervised methods also in our investigations, where there exists a trade-off between unnatural deformations and the method not learning to align anatomical structures, resulting again in a "zebra crossing" pattern.

\subsection{stage$1$ vs. stage$2$}

We want to investigate whether there is true benefit in using our multi-stage architecture over only a single stage.
With $\mathcal{W}$, we show significant reduction in negJAC of all stage$2$ models over their stage$1$ counterparts (compare subscript $1$ against $12$ models in Table \ref{tab:results splits}), suggesting that stage$2$ architectures are indeed better at preventing folding. All stage$2$ models show small but significant increase in MSE over their stage$1$ counterparts. Rib-based stage$2$ models $RC_{12}$ and $R_{12}$ show significant increase in DCR over their stage$1$ counterparts, while the increase in DCL in $L_{12}$ is not significant compared to $L_{1}$. We conclude that we find statistical evidence that our multi-stage architecture is superior to the single stage architecture in terms of preventing folding and that rib-based methods particularly benefit from the second stage also in terms of anatomical overlap, as also illustrated in Figure \ref{fig:stages comparison}. 
Visually, this results in better alignment of the penalized anatomical structures, which, for rib-based methods causes the "zebra crossing" pattern to disappear and to make fine details better visible. For lung-based methods, this results in better alignment of lung and heart boundaries, which comes at the cost of losing information on pathological differences in lung and heart shape.

\subsection{Inter- vs. intra-patient training}

We observed that for the task of registering images from the same patient, intra-patient data is beneficial over inter-patient data (around $5\%$ in terms of DCR, cf. Table \ref{tab:intra vs inter}), when training with the same amount of moving images but pairing it with a random image from another patient as opposed to the same patient. Through $\mathcal{W}$, we find significant differences between intra- and inter-patient training in terms of higher MSE but lower negJAC. Difference in SSIM was not found to be significant. $\mathcal{W}$ shows significant differences between intra- and inter-patient training in terms of higher DCR. No significant differences in DCL were found, meaning that there are differences between these two training schemes, that can not be explained by analyzing lung overlap alone.
We therefore suggest to at least fine tune models on intra-patient data if they are intended to be used for this task.

\subsection{Hole dropout and weak labeling}

We investigate whether the higher DCR scores achieved through hole dropout and weak labeling reported in Section \ref{sec:scores} are statistically significant compared to the standard iterative segmentation approach. With $\mathcal{F}$ we find statistical differences between methods that are also pairwise significant for all combinations according to $\mathcal{N}$. This supports our hypothesis that modeling imperfect annotations with holes as instance memory improves segmentation results and incorporating weak labels by models trained on only $17$ images significantly boosts segmentation performance.

\subsection{Impact on visualization of differences}

The visual impact of our findings can be seen in Figures \ref{fig:lung rib diff}, \ref{fig:result comparison} and \ref{fig:stages comparison}, where we illustrate prevalent problems of SOTA methods on examples from the NIH data set. If ribs are not overlapping, a "zebra crossing" pattern appears in the difference image, where dark and light stripes are caused by ribs being in different positions between warped and fixed image. This is not a pathological difference, but it is impacting the visibility of small pathological and non-pathological details, like nodules or foreign objects. All SOTA methods investigated, lung penalized as well as unsupervised, show this pattern, as ribs seem to be a weaker signal for guidance of the unsupervised optimization loss parts than for example diaphragm boundaries. Rib pairs penalization, particularly when employed in our multi-stage framework, significantly reduces these patterns. 

When lungs are forced to overlap, boundaries of the lung fields and the heart become aligned, often causing the warp field to collapse, which is also measurable by negJAC. This obfuscates changes when a pathology manifests itself in a deformed heart contour (like for example cardiomegaly) or in a deformed lung contour (like effusion or consolidation). 

When not forcing these boundaries to overlap, differences in breathing patterns become visible.

\subsection{Limitations}
\label{sec:limitations}

Since we use ribs as guidance during registration, problems could occur when ribs are broken or not visible. We observed on several cases that the regularization power of the network counters this, but further investigation is needed. We also occasionally encountered a miscounting of ribs but are confident, that this problem can be solved by more expert labeled data for the first rib pair. We also noted that clavicles are highlighted as very prominent differences in the images. This could be countered by bone removal algorithms for clavicles but we point out that registration frameworks used in combination with bone removal algorithms usually perform lung-based registration beforehand and therefore would run into the exact problems our methods are preventing. 
Additionally, non-pathological breathing patterns are shown as differences with our method. In the future, it would be interesting to investigate, whether a combination of different penalization methods in one image is able to show only relevant changes. 

\section{Summary and Conclusion}
\label{sec:conclusion}

In this paper, we analyzed existing biases of chest X-ray registration methods and their evaluation. We proposed a new method for chest X-ray registration through rib pairs supervision capable of handling large and subtle changes at the same time.  

For the registration approach, we built on the VectorCNN architecture from \cite{Mansilla2020}, which we extended to a multi-stage architecture, in order to be able to attend to the fine details of rib contours. The second stage directly takes the lower resolution displacement field and the input image and acts as a refinement stage for the warp field, improving anatomic registration quality. With this method the rib pairs works as a natural grid during training, significantly reducing unnatural deformations of the warp field to $1/6$ of the SOTA while at the same time increasing the overlap of ribs by more than $25\%$. For segmentation, we built upon the idea of instance memory originally introduced for vertebrae by \cite{Lessmann2019} and employed by \cite{Wessel2019} for individual ribs, which we applied to pairs of ribs, by informing the segmentation of the next rib pair with the output of a network segmenting the previous rib pair. Coupling this with a region-dropout technique simulating imperfect segmentation masks, together with rule based weak label correction, we make the approach more robust to disconnected regions in the segmentation output. In this way, we improve the segmentation performance significantly over using pure instance memory and strong labels only, making the method applicable to generate weak labels for the large unlabeled NIH data set. 

To contrast our findings with the literature, we summarized explicit or implicit assumptions on lung overlap during training or evaluation of deep learning based chest X-ray registration methods in Table \ref{tab:related work} and evaluated our method against the SOTA in weakly supervised and unsupervised chest X-ray registration.
While it is to be expected, that models optimized for different anatomy overlap and selected through hyper parameter search as being the best for this particular task shine when assessing with metrics measuring this overlap, it is still surprising how little the overlap of ribs, which is a clearly visible structure in most chest X-ray images, is considered by methods not employing a rib based penalty. The best SOTA method \cite{Mansilla2020} achieves average dice overlap of rib pairs of around $0.5$, while at the same time achieving lung overlap scores above $0.95$, outperforming other methods in this task. Additionally, the lung and unsupervised methods have been shown to have a clear tendency to collapse warp fields. This fact can not be accurately measured by metrics like DCL or pixel similarity metrics, which are also not capable of measuring the effect of intra-patient vs. inter-patient training strategies, demonstrating again the benefit of observing rib overlap for evaluation of chest X-ray registration. 
 
Our findings are particularly relevant when large changes occur between follow up studies, where the SOTA methods force the alignment of lung boundaries or the similarity of pixel values.  
The explained phenomena impact the visualization of the difference image in two ways, as also shown in Figure \ref{fig:lung rib diff}, \ref{fig:result comparison} and \ref{fig:stages comparison}: Non-overlapping ribs cause "zebra crossing" patterns of alternating ribs between the warped and the fixed image to appear as very prominent differences, potentially obfuscating more subtle differences like nodules or small foreign objects like catheters. Additionally, forced alignment of lung boundaries hides potential differences between lung and heart shape, which makes pathological changes like cardiomegaly, effusion or consolidation disappear. In the future, we plan to evaluate the clinical benefit of our method on improved confidence in chest X-ray readings in a user study with radiologists.









\printcredits

\section*{Declaration of competing interest}

The authors declare that they have no known competing financial interests or personal relationships that could have appeared to influence the work reported in this paper.

\section*{Acknowledgements}

VRVis is funded by BMK, BMDW, Styria, SFG, Tyrol and Vienna Business Agency in the scope of COMET - Competence Centers for Excellent Technologies (879730) which is managed by FFG. Thanks go to our project partner Agfa Radiology Solutions for valuable input.

\bibliographystyle{cas-model2-names}

\bibliography{references}

%

\end{document}